\documentclass[a4paper, amsfonts, amssymb, amsmath, reprint, showkeys, footinbib, ,twoside,superscriptaddress,floatfix,longbibliography]{revtex4-1}

\usepackage{graphicx} %
\usepackage{amsmath}
\usepackage{comment}
\usepackage{booktabs}
\usepackage{hyperref}
\usepackage{url}
\usepackage{adjustbox}
\usepackage{array}
\usepackage{xcolor}

\newcommand{\rev}[1]{{\color{black} #1 }}

\newcommand{\figLabelCapt}[1]{\textbf{\MakeLowercase{{#1}}}}
\newcommand{\refSub}[2]{\hyperref[#2]{\ref{#2}\figLabelCapt{#1}}}

\begin{document}

\title{Latent Ewald summation for machine learning of long-range interactions}

\author{Bingqing Cheng}
\email{bingqingcheng@berkeley.edu}
\affiliation{Department of Chemistry, University of California, Berkeley, CA, USA}
\affiliation{The Institute of Science and Technology Austria, Am Campus 1, 3400 Klosterneuburg, Austria}

\date{\today}

\begin{abstract}
Machine learning interatomic potentials (MLIPs) often neglect long-range interactions, such as electrostatic and dispersion forces. In this work, we introduce a straightforward and efficient method to account for long-range interactions by learning a hidden variable from local atomic descriptors and applying an Ewald summation to this variable. We demonstrate that in systems including \rev{charged and polar} molecular dimers, bulk water, and water-vapor interface, standard short-ranged MLIPs can lead to unphysical predictions even when employing message passing. The long-range models effectively eliminate these artifacts, with only about twice the computational cost of short-range MLIPs.

\end{abstract}

\maketitle

%\section{Introduction}

Machine learning interatomic potentials (MLIPs) can learn from reference quantum mechanical calculations and then predict the energy and forces of atomic configurations quickly, allowing for a more accurate and comprehensive exploration of material and molecular properties at scale~\cite{keith2021combining,unke2021machine}. Most state-of-the-art MLIP methods use a short-range approximation: the effective potential energy surface experienced by one atom is determined by its atomic neighborhood. This approximation implies that the total energy is the sum of atomic contributions, which also makes the MLIPs scale linearly with system size.

The short-range MLIPs, however, neglect all kinds of long-range interactions such as Coulomb and dispersion.
Although short-range potentials may be sufficient to describe most properties of homogeneous bulk systems~\cite{yue2021short},
they may fail for liquid-vapor interfaces~\cite{niblett2021learning},
dielectric response~\cite{rodgers2008interplay,cox2020dielectric},
dilute ionic solutions with Debye-H\"{u}ckel screening,
and interactions between gas phase molecules~\cite{huguenin2023physics}.

There has been a continuous effort to
incorporate long-range interactions into MLIPs.
One can include empirical electrostatics and dispersion baseline corrections~\cite{niblett2021learning,lee2022high,unke2021spookynet}, 
but for many systems, such baseline is not readily available.
Another option is to predict effective partial charges to each atom, which are then used to calculate long-range electrostatics~\cite{unke2019physnet,ko2021fourth,gao2022self,sifain2018discovering,gong2024bamboo,shaidu2024incorporating}.
For example, the fourth-generation high-dimensional neural network potential (4G-HDNNPs)~\cite{ko2021fourth} predicts the electronegativities of each nucleus, and then use a charge equilibration scheme~\cite{rappe1991charge} to assign the charges. 
4G-HDNNPs are trained directly to reproduce atomic partial charges from reference quantum mechanical calculations, although partial charges are not physical observable and their values depend on the specific partitioning scheme used~\cite{sifain2018discovering}.
In a similar vein, the deep potential long-range (DPLR)~\cite{zhang2022deep} learns maximally localized Wannier function centers (MLWFCs) for insulating systems, and
the self-consistent field neural network (SCFNN)~\cite{gao2022self} predicts the electronic response via the position of the MLWFCs.
Message passing neural networks (MPNNs)~\cite{schutt2017schnet,batzner20223,deng2023chgnet,haghighatlari2022newtonnet} 
employ a number of graph convolution layers to communicate information between atoms, thus capturing long-range interaction up to the local cutoff radius times the number of layers.
However, if parts of the system are disconnected on the graph, e.g. two molecules with a distance beyond the cutoff, the message passing scheme does not help.
\rev{Another class of methods is to learn the long-range  descriptors and interactions in the reciprocal space with learnable frequency filters~\cite{yu2022capturing,kosmala2023ewald}.}
Finally, a very interesting approach is the long-distance equivariant (LODE) method~\cite{grisafi2019incorporating,huguenin2023physics},
which uses local descriptors to encode the
Coulomb and other asymptotic decaying
potentials ($1/r^p$) around the atoms,
\rev{and a related, density-based long-range descriptors~\cite{faller2024density}.}

Here, we propose a simple method, the Latent Ewald Summation (LES), for accounting for long-range interactions of atomistic systems.
The method is general and can be incorporated into most existing MLIP architectures, 
including potentials based on local atomic environments (e.g. HDNNP~\cite{behler2007generalized}, Gaussian Approximation Potentials (GAP)~\cite{bartok2010gaussian}, Moment Tensor Potentials (MTPs)~\cite{shapeev2016moment}, atomic cluster expansion (ACE)~\cite{drautz2019atomic}) and MPNN (e.g.
NequIP~\cite{batzner20223}, MACE~\cite{batatia2022mace}).
In the present work, we combine LES with
Cartesian atomic cluster expansion (CACE) MLIP~\cite{cheng2024cartesian}.
After describing the algorithm, we benchmark LES on selected molecular and material systems.

\section{Results}

\subsection{Theory}
For a periodic atomic system,
the total potential energy is decomposed into a short-range and a long-range part,
i.e. $E = E^{sr} + E^{lr}$.
As is standard in most MLIPs, the short-range energy is summed over the atomic contribution of each atom $i$,
\begin{equation}
    E^{sr} = \sum_i E_{\theta}(B_i),
\end{equation}
where $E_{\theta}$ is a multilayer perceptron with parameters $\theta$ that maps
the invariant features ($B$) of an atom to its short-range atomic energy.
$B$ can be any invariant features used in different MLIP methods, including those based on local atomic environment descriptors such as ACE~\cite{drautz2019atomic}, atom-centered symmetry functions~\cite{behler2007generalized}, smooth overlap of atomic positions (SOAP)~\cite{bartok2013representing}, or any latent invariant features in MPNNs.

For the long-range part, another multilayer perceptron with parameters $\phi$ maps the invariant features of each atom $i$ to a hidden variable, i.e.
\begin{equation}
    q_i = Q_{\phi}(B_i).
    \label{eq:q}
\end{equation}
The structure factor $S(\mathbf{k})$ of the hidden variable is defined as
\begin{equation}
    S(\mathbf{k}) = \sum_i q_i e^{i\mathbf{k}\mathbf{r}_i},
\end{equation}
where $\mathbf{k} = (2\pi n_x/L_x, 2\pi n_y/L_y, 2\pi n_z/L_z)$ is a reciprocal vector of the orthorhombic cell, 
and $\mathbf{r}_i$ is the Cartesian coordinates of atom $i$.
The long-range energy is then obtained \rev{using an Ewald summation form that best captures the electrostatic potential ($1/r$)~\cite{williams1971accelerated}}:
\begin{equation}
    E^{lr} = \dfrac{1}{V} \sum_{0<k<k_{c}} \dfrac{e^{-\sigma^2 k^2/2}}{k^2} |S(\mathbf{k})|^2,
    \label{eq:ewald}
\end{equation}
where $\sigma$ is a smearing factor which we typically set to 1~\AA{} \rev{with justifications in the Methods},
$k = |\mathbf{k}|$ is the magnitude,
and $k_c$ is the maximum cutoff.
$q$ can be multi-dimensional, in which case the total long-range energy is aggregated over contributions from different dimensions of $q$ after the Ewald summation.

The LES method can be interpreted in two ways. First, the hidden variable $q$ is analogous to the environmental-dependent partial charges on each atom. This implies that the method is at least as expressive as those explicitly based on learning partial atomic charges, as it would yield the same results if $q$ replicated the partial charges. Additionally, $q$ can be multidimensional, potentially enhancing expressiveness further. Unlike partial charges, $q$ is not constrained by requirements such as charge neutrality or correct charge magnitudes. 
\rev{As the Ewald summation in Eqn.~\eqref{eq:ewald} omits the $k=0$ term,
a non-zero net $q$ does not cause energy divergence issues.}
Since the Ewald summation of $q$ doesn’t need to correspond to a physical electrostatic potential, Eqn.~\eqref{eq:ewald} omits the self-interaction term present in the regular Ewald summation for long-range charge interactions.
The second interpretation of LES is as a mechanism that allows atoms far apart in the simulation box to communicate their local information. In this sense, LES is related to the recent Ewald-based long-range message passing method, which facilitates message exchange between atoms in reciprocal space~\cite{kosmala2023ewald}.

\subsection{Example on molecular dimers}

\begin{figure*}
  \centering
      \vspace{-0.5cm}
      \includegraphics[width=0.8\textwidth]{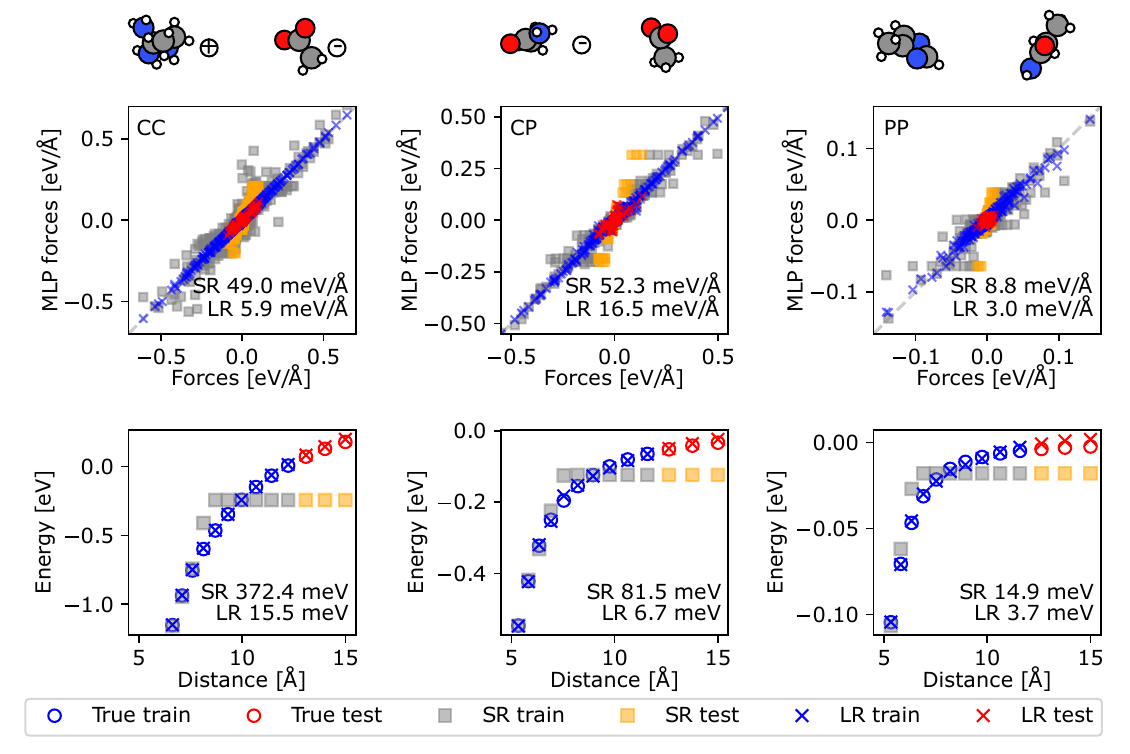}
    \caption{
Comparison of the short-range (SR) and long-range (LR) machine learning interatomic potential performance for \rev{three} dimer classes: charged-charged (CC), charged-polar (CP), and polar-polar (PP).
For each class,
the upper panel shows a snapshot of the system with the charge states indicated,
the middle panel shows the parity plot for the force components,
and the lower panel shows the binding energy curve, which is the potential energy difference between the dimer, and two isolated \rev{and relaxed} monomers. 
The root mean square errors (RMSE) for the energy and force components of the test sets are shown in the insets.
    }
    \label{fig:dimer}
\end{figure*}

We benchmark the LES method on the binding curves between dimers of \rev{charged (C) and polar (P) molecules} at various separations in a periodic cubic box with a 30~\AA{} edge length. The dataset~\cite{huguenin2023physics}, \rev{originally from the BioFragment Database~\cite{burns2017biofragment},} includes energy and force information calculated using the HSE06 hybrid density functional theory (DFT) with a many-body dispersion correction. We selected one example from each of the \rev{three dimer classes (CC, CP, PP)}, derived from the combination of the three monomer categories. Figure~\ref{fig:dimer} shows a snapshot of each example.

The goal of this benchmark is to evaluate whether the MLIP models can extrapolate dimer interactions at larger separations based on training data from smaller distances.
For each molecular pair, the training set consists of 10 configurations with dimer separation distances between approximately 5~\AA{} and 12~\AA{}, and the test set includes 3 configurations with separations between approximately 12~\AA{} and 15~\AA. 

For each molecular pair, we trained a short-range (SR) model using CACE with a cutoff $r_\mathrm{cut} = 5$~\AA, 6 Bessel radial functions, $c = 8$, $l_\mathrm{max} = 2$, $\nu_\mathrm{max} = 2$, $N_\mathrm{embedding} = 3$, and one message passing layer ($T = 1$). 
It should be noted that this setting of the ``SR'' model already achieves a perceptive field of 10~\AA{} through the message passing layer, which is quite typical for current MPNNs. 
In comparison, more traditional MLIPs based on local atomic descriptors typically use a cutoff of around 5 or 6~\AA{}, making them even more short-ranged.

In the long-range (LR) model, the short-range component $E^{sr}$ used the same CACE setup, while the long-range component $E^{lr}$ employed a 4-dimensional hidden variable computed from the same CACE $B$-features and utilized Ewald summation (Eqn.~\eqref{eq:ewald}) with $\sigma = 1$~\AA{} and a $\mathbf{k}$-point cutoff of $k_c = 2\pi/3$. It is important to fit both energy and forces for these datasets, as fitting only to a few energy values may result in models that accurately predict binding energy but perform poorly on forces.

Fig.~\ref{fig:dimer} shows that the LR MLIPs outperform the SR MLIPs in all cases. 
Furthermore, SR models fail to adequately capture the training data, as indicated by flattening of the binding curves and the large discrepancies between some predicted and true forces (gray symbols in Fig.~\ref{fig:dimer}). 
The primary limitation of the SR models is that molecules separated by distances beyond the MLIP cutoff exist on independent atomic graphs, rendering the message-passing layers ineffective for communication between the dimers.
Direct comparison of the accuracy of the present LR models with the LODE method~\cite{huguenin2023physics} is challenging, as different training protocols were used in Ref.~\cite{huguenin2023physics}, and LODE depends significantly on the choice of the potential exponent $p$ for each dimer class. 
However, based solely on RMSE values, the LR models presented here appear to be more accurate. For example, in the CC class, the energy RMSE is 15.5 meV, compared to approximately 0.1 eV for LODE with the optimal choice of $p$.

\subsection{Example on molten NaCl}
\rev{
Molten bulk sodium chloride (NaCl) presents non-negligible long-range electrostatic interactions. 
The dataset from Ref.~\cite{faller2024density} contains 1014 structures (80\% train and 20\% validation) of 64 Na and 64 Cl atoms. 
Table~\ref{tab:res_nacl} compares the root-mean-square percentage error (RMSPE) using different methods.
$r_{\mathrm{cut}}=~$6 \AA{} were used for all models except for the LODE flexible (7~\AA).
The CACE-LR models use 6 Bessel radial functions, $c = 12$, $l_\mathrm{max} = 3$, $\nu_\mathrm{max} = 3$, $N_\mathrm{embedding} = 3$, zero ($T=0$) or one message passing layer ($T = 1$), a 4-dimensional $q$,
$\sigma = 1$~\AA{} and a $\mathbf{k}$-point cutoff of $k_c = 2\pi/3$ in Ewald summation.
The details for the other models are in Ref.~\cite{faller2024density}.

\begin{table}[h]
\caption{\label{tab:res_nacl}%
\rev{The root-mean-square percentage error (RMSPE) for the validation set of molten NaCl, using the short-ranged SOAP descriptor~\cite{bartok2010gaussian,bartok2013representing,jinnouchi2019fly}, two setting of LODE~\cite{grisafi2019incorporating,huguenin2023physics} descriptors (minimal and flexible),  MACE~\cite{batatia2022mace} with zero ($T=0$) or one message passing layers ($T=1$) as well as using invariant and equivariant messages, a density-based long-range model (Density-LR)~\cite{faller2024density}, and CACE-LR with zero ($T=0$) or one message passing layers ($T=1$).
The fits other than the CACE models are from Ref.~\cite{faller2024density}.}
}

\begin{ruledtabular}
\begin{tabular}{lcc}
 &RMSPE E (\%)& RMSPE F (\%)\\
 \hline
 SOAP~\cite{bartok2010gaussian,bartok2013representing,jinnouchi2019fly} &7.7 & 12.6\\
 MACE~\cite{batatia2022mace} T=0 & 8.5 & 8.5\\
 MACE~\cite{batatia2022mace} T=1 invariant & 3.0 & 3.1\\
 MACE~\cite{batatia2022mace} T=1 equivariant & 2.1 & 2.2\\
 LODE~\cite{grisafi2019incorporating,huguenin2023physics} minimal & 5.6 & 3.7\\
 LODE~\cite{grisafi2019incorporating,huguenin2023physics} flexible & 7.0 & 11.0 \\
 Density-LR~\cite{faller2024density} &2.2&3.3\\
 \textbf{CACE-LR T=0} & \textbf{1.9} & \textbf{2.3} \\ 
\textbf{CACE-LR T=1} & \textbf{1.4} & \textbf{1.8} \\
 \end{tabular}
\end{ruledtabular}
\end{table}

Table~\ref{tab:res_nacl} shows that the short-ranged models, including SOAP~\cite{bartok2010gaussian,bartok2013representing,jinnouchi2019fly} and MACE~\cite{batatia2022mace} $T=0$, have the worst performances.
All the long-ranged models, including LODE~\cite{grisafi2019incorporating,huguenin2023physics} with different settings, a density-based long-range model~\cite{faller2024density},
and CACE-LR outperform the purely SR methods.
MACE~\cite{batatia2022mace} with message passing ($T=1$), especially the one using equivariant features, also achieves accurate results,
as message passing increases the effective perceptive field.
Overall, CACE-LR models, with and without message passing ($T=0$ and $T=1$), obtain the best accuracy.
}

\subsection{Example on bulk water}

As an example application to dipolar fluids, we applied the LES method to a dataset of 1,593 liquid water configurations, each containing 64 molecules~\cite{cheng2019ab}. 
The dataset was calculated using revPBE0-D3 DFT. 
For the CACE representation, we used a cutoff of $r_\mathrm{cut} = 5.5$~\AA, 6 Bessel radial functions with $c = 12$, $l_\mathrm{max} = 3$, $\nu_\mathrm{max} = 3$, $N_\mathrm{embedding} = 3$, and no message passing ($T = 0$) or no message passing layer ($T = 1$). 
For the long-range component, we used a 4-dimensional $q$, a maximum cutoff of $k_c = \pi$, and $\sigma = 1$~\AA{} in the Ewald summation.

\begin{figure}
  \centering
  \includegraphics[width=0.45\textwidth]{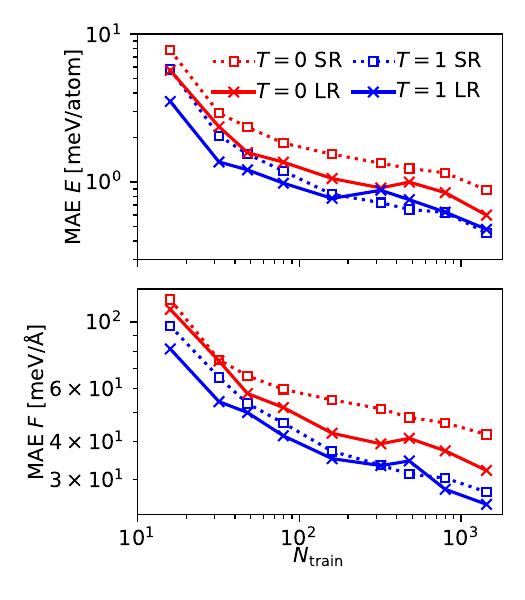}
    \caption{
Learning curves of energy ($E$) and force ($F$) mean absolute errors (MAEs) on the bulk water dataset~\cite{cheng2019ab},
using short-range (SR) or long-range (LR) models and with or without  massage passing layers ($T=0$ or $T=1$).
    }
    \label{fig:water-lr}
\end{figure}

The learning curves in Fig.~\ref{fig:water-lr} demonstrate that message passing significantly improves the accuracy of MLIPs, consistent with previous studies~\cite{batzner20223,batatia2022mace}. 
The LR component further reduces the error for both models without a message passing layer ($T=0$) and with a message passing layer ($T=1$). 
The improvement is particularly notable in the $T=0$ scenario. 
For the $T=1$ models, the LR component results in a smaller reduction in errors, probably because the $T=1$ SR models already capture atomic interactions up to 11~\AA. 
Nevertheless, the $T=1$ LR model demonstrates greater efficiency in learning with fewer data. 
We also compared the inference speeds of these models during molecular dynamics (MD) simulations, as detailed in the Methods section.

\begin{figure}[h]
  \centering
\includegraphics[width=0.45\textwidth]{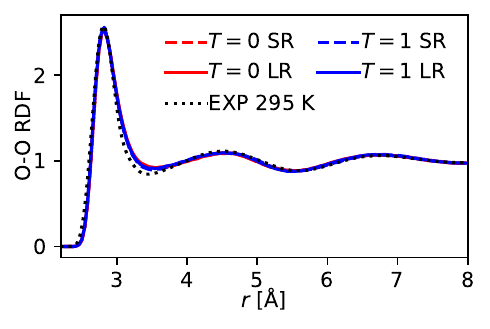}
    \caption{
    Predicted oxygen-oxygen radial distribution functions (RDF) of water at 300~K and 1~g/mL,
    using short-range (SR) or long-range (LR) models and with or without massage passing layers ($T=0$ or $T=1$).
    The experimental O-O RDF at ambient conditions was obtained from Ref~\cite{skinner2014structure}.
    }
    \label{fig:water-gor}
\end{figure}

To investigate how message passing and long-range interactions affect predicted structural properties, we performed MD simulations of bulk water with a density of 1~g/mL at 300~K using each model. 
The upper panel of Fig.~\ref{fig:water-gor} shows the oxygen-oxygen (O-O) radial distribution function (RDF) computed with different MLIPs. 
All computed O-O RDFs are indistinguishable and in excellent agreement with the experimental results of the X-ray diffraction measurements~\cite{skinner2014structure}. 
This suggests that local representations are sufficient to accurately predict the RDFs of bulk liquid water, consistent with previous findings~\cite{yue2021short}.

\begin{figure}[h]
  \centering
\includegraphics[width=0.45\textwidth]{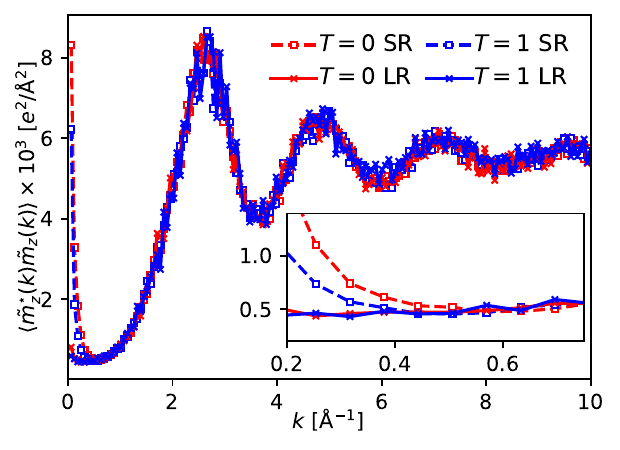}
    \caption{
Predicted dipole density correlations in reciprocal space, using short-range (SR) or long-range (LR) models and with or without massage passing layers ($T=0$ or $T=1$).
The inset shows a zoom-in of the small $k$ values.
    }
    \label{fig:water-dc}
\end{figure}

We then investigated the effect of long-range interactions on the dielectric properties of water by computing the longitudinal component of the dipole density correlation function~\cite{cox2020dielectric}, $\langle \Tilde{m}_z^{\star}(k) \Tilde{m}_z(k)\rangle$, where $\Tilde{m}$ is the Fourier transform of the molecular dipole density and $\mathbf{k} = k \mathbf{z}$ is along the z-axis. The dipole correlation function at the long-wavelength limit can be used to determine the dielectric constant of water~\cite{gao2022self}.

As shown in Fig.~\ref{fig:water-dc}, while the dipole density correlation functions predicted by different MLIPs are in excellent agreement for most values of $k$, discrepancies emerge at long wavelengths. Specifically, the results from short-ranged models sharply increase as $k \rightarrow 0$. This divergence has also been observed in previous studies comparing the SR and LR models~\cite{cox2020dielectric,gao2022self}. The divergence of the $T=0$ short-ranged MLIP appears at a $k$ value corresponding to a real length scale of approximately 11~\AA{}, while the $T=1$ SR model diverges at about 16~\AA{}.
Interestingly, these length scales, where divergence occurs, exceed the effective cutoff radius, suggesting that SR MLIPs may partially describe long-range effects beyond their cutoffs in a mean-field manner. When comparing the two SR MLIPs, the message passing layer delays the onset of divergence at small $k$ values but does not eliminate it. In this scenario, only true LR models can adequately describe the dielectric response.

\begin{figure}
  \centering
  \includegraphics[width=0.45\textwidth]{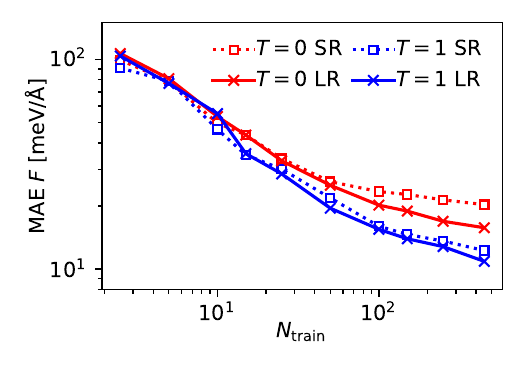}
    \caption{
Learning curves on the liquid–vapor water interface dataset~\cite{niblett2021learning} using the short-range (SR) and long-range (LR) models with no message passing ($T=0$) or one message passing layers ($T=1$).
    }
    \label{fig:slab-lr}
\end{figure}

\subsection{Example on interfacial water}

It is widely recognized that short-range MLIPs fall short in describing interfaces~\cite{niblett2021learning, gao2022self}. 
To demonstrate the efficacy of the LES method for interfaces, we used a liquid-water interface dataset from Ref.~\cite{niblett2021learning}, computed with the revPBE-D3 functional. 
We selected a subset of 500 liquid-vapor interface configurations, each containing 522 water molecules, far fewer than the approximately 17,500 training configurations used in Ref.~\cite{niblett2021learning} to train the DeePMD model~\cite{zhang2022deep},
which may be attributed to the learning efficiency of CACE.

We employed the same settings for fitting the MLIPs as used for the bulk water dataset. 
The learning curves for forces are shown in Fig.~\ref{fig:slab-lr}. 
The energy errors are very low, reaching less than 0.1~meV/atom in mean absolute error (MAE) for all models using only 50 training configurations. 
The force errors are also small, ranging from about 14~meV/\AA{} to 27~meV/\AA{} in MAE when trained on 90\% of the dataset. 
The learning behavior is similar to that observed in the bulk water case: both message passing and long-range interactions enhance the accuracy of the MLIPs.

\begin{figure*}
  \centering
\includegraphics[width=0.75\textwidth]{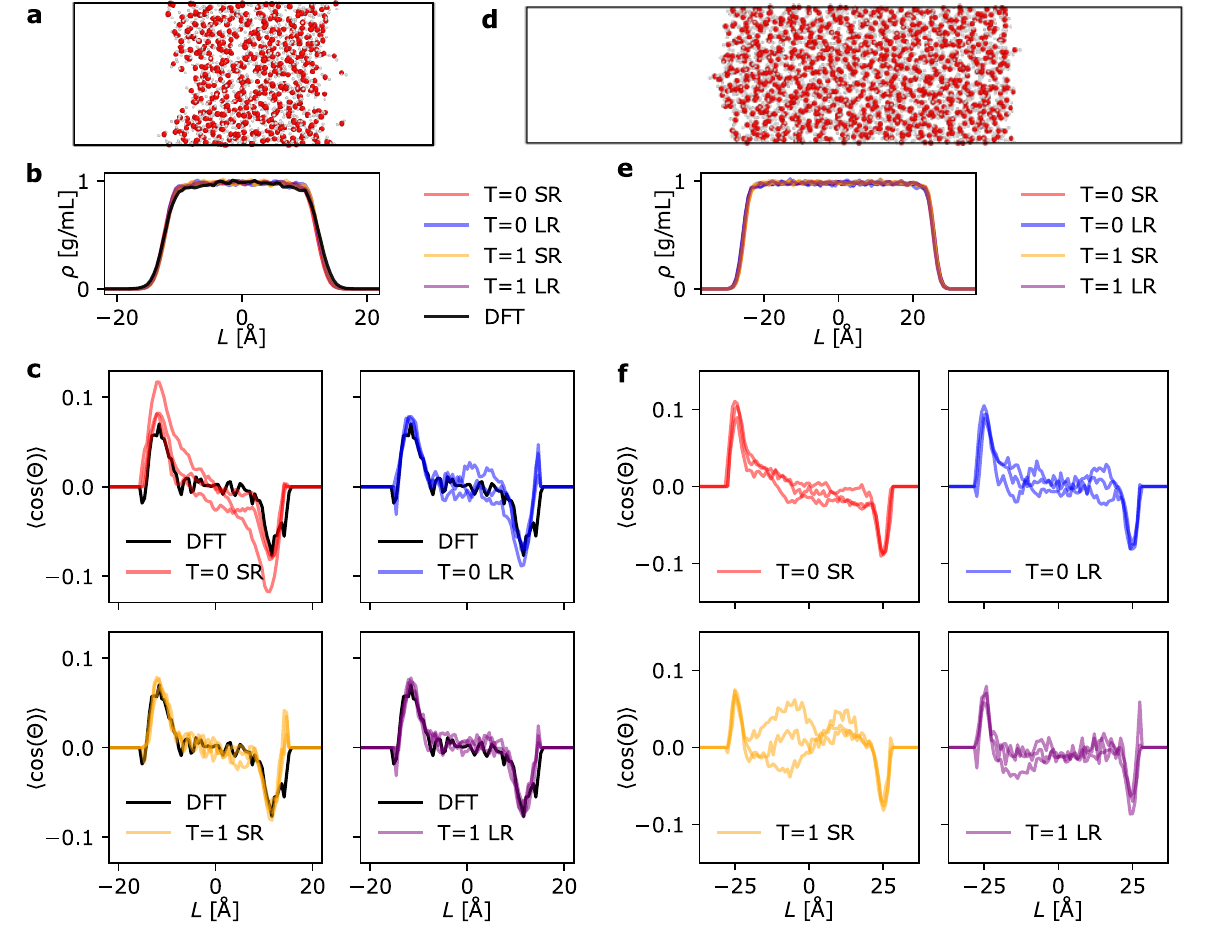}
    \caption{
A comparison of the interface properties of water predicted using short-range (SR) or long-range (LR) models and with or without massage passing layers ($T=0$ or $T=1$). 
\textbf{a} shows a snapshot of the thinner water slab configuration,
\textbf{b} shows the water density profile, and \textbf{c} shows average cosine (c) the average $\cos(\theta)$ for the angle formed by the water dipole moment and z-axis.
DFT results are from Ref.~\cite{niblett2021learning}.
\textbf{d}, \textbf{e} and \textbf{f} show the snapshot, the
water density, and $\langle\cos(\theta)\rangle$ for the thicker water slab, respectively.
    }
    \label{fig:waterslab-properties}
\end{figure*}

For each of the $T=0$ SR, $T=0$ LR, $T=1$ SR, and $T=1$ LR models, we trained three MLIPs using different random splits of 90\% for training and 10\% for testing. 
We then simulated the liquid–vapor interface at 300~K using both a thinner slab (Fig.~\ref{fig:waterslab-properties}a) and a thicker slab of about double the thickness (Fig.~\ref{fig:waterslab-properties}d). 
The resulting density profiles of the water slabs are shown in Fig.~\ref{fig:waterslab-properties}b,e. 
All SR and LR models predict similar density profiles, and for the thinner slab, all models accurately reproduce the density profiles of the reference DFT water data.

In addition to density profiles, we evaluated the orientational order profiles, shown in Fig.~\ref{fig:waterslab-properties}c,f, 
by computing the angle, $\theta$, between the dipole orientation of water molecules and the $z$-axis. 
For each model, the three different fits of MLIPs provide slightly different predictions, \rev{and the variances are larger for the SR models compared to the LR models.}
At the interfaces, water molecules tend to form a dipole layer, which is screened by subsequent layers, ensuring that the bulk does not exhibit net dipole moments~\cite{niblett2021learning}. 
However, without long-range interactions, this screening effect is not adequately captured, leading to extended dipole ordering into the bulk as seen in the $T=0$ SR model.
The introduction of message passing ($T=1$ SR model) alleviates this issue at least for the thinner slab, but it also introduces an artifact: dipole ordering in the bulk along the opposite direction for the thicker slab. 
In contrast, \rev{on average and with smaller variance between different fits}, the LR models recover the correct orientational order profiles, effectively capturing the screening effect and accurately representing the physical polarization behavior at the liquid-vapor interfaces.

\section{Discussion}

\rev{
The current LES method can be extended in a number of ways.
The form of the Ewald summation in Eqn.~\eqref{eq:ewald} was selected to best capture the electrostatic potential ($1/r$), which dominates long-range interactions. 
It remains unclear what is the effect of not directly accounting for other decaying potentials.
Different forms can be used to capture other decaying potentials $1/r^p$ with different $p$ components.
For examples,
to best capture the London dispersion ($1/r^6$), one can use~\cite{chen1997fast, williams1971accelerated}.
\begin{multline}
       E_6^{lr} =
\left( \frac{\pi^{3/2}}{24 V} \right) \sum_{k>0} 
k^3 \\
\left[ \pi^{1/2} \mathrm{erfc} (b) + \left( \frac{1}{2b^3} - \frac{1}{b} \right) \times \exp(-b^2) \right] |S(\mathbf{k})|^2 ,
\end{multline}
where $b^2 = \sigma^2 k^2/2$, and $\mathrm{erfc}$ denotes the complimentary error function. 
Currently, the hidden variable $q$ is rotational invariant, and it may be extended to adopt a vector or tensor form to capture the dipole and multiple moments on an atom. 
}

\rev{The relation between LES and the LODE method~\cite{grisafi2019incorporating,huguenin2023physics} is worth exploring.
The key difference is the ordering of local and global interactions.
In LES, the hidden variable $q_i$ is determined by the local atomic environment-dependent features $B_i$ of atom $i$, reflecting the nearsightedness principle: partial atomic charges depend mainly on local environments, similar to the short-range approximation in standard MLIPs.
Although it should be pointed out that such locality approximation remains empirical rather than systematic.
These local $q$ then interact globally via the Ewald summation to determine the global long-range energy.
In contrast, in LODE~\cite{grisafi2019incorporating,huguenin2023physics} the potential field (e.g. electrostatic potential and other $1/r^p$ decaying potentials) generated by all the atoms in the system is calculated in the reciprocal space via Ewald summation, and such field near a central atom $i$ up to some cutoff radius is then projected onto a set of basis functions to form the LODE descriptors. 
Finally, the short-range descriptors and the LODE descriptors of atom $i$ are used to predict its energy.
The density-based long-range descriptors~\cite{faller2024density} are similar to LODE, but the global atomic density itself is used instead of the field.
In summary, LES uses local descriptors to predict global long-range energy, while LODE uses global field to inform local descriptors. 
There may be underlying mathematical connections between the LES and LODE, but this is not obvious, particularly considering the non-linearity introduced by the multilayer perceptron in LES that predicts $q_i$ from $B_i$ (Eqn.~\eqref{eq:q}).

LES also shares similarities with the recently proposed Ewald-based long-range message-passing method~\cite{kosmala2023ewald}, as both utilize structure factors. While LES directly determines the long-range energy from the structure factor of a hidden low-dimensional variable (Eqn.\eqref{eq:ewald}), the Ewald message-passing method~\cite{kosmala2023ewald} employs a learnable frequency filter to map the structure factor of atomic descriptors to real space. This mapping generates a long-range message for each atom that is used to update atomic descriptors during the message passing step.
LES can thus be interpreted as a minimalist version of Ewald message-passing,
with only one long-range message passing layer, compressed message dimension,  fixed frequency filter, and simplified readout.
Such minimalist design makes the model light-weight, and allows for easier physical interpretation.
}

%\section{Conclusions}

In summary, we present a simple and general method, the Latent Ewald Summation (LES), for incorporating long-range interactions in atomistic systems. 
Unlike many existing LR methods, LES does not require user-defined electrostatic or dispersion baseline corrections~\cite{niblett2021learning,lee2022high,unke2021spookynet}, does not rely on partial charges or Wannier centers during training~\cite{unke2019physnet,ko2021fourth,gao2022self,sifain2018discovering,gong2024bamboo,shaidu2024incorporating}, and does not utilize charge equilibration schemes~\cite{rappe1991charge}. 
\rev{Moreover, LES shows best accuracy compared to other long-range method and message passing neutral networks in the example on molten salt.}

Our results demonstrate that the LR model effectively reproduces correct dimer binding curves for pairs of charged and polar molecules, long-range dipole correlations in liquid water, and dielectric screening at liquid-vapor interfaces. 
In contrast, standard short-ranged MLIPs, even when utilizing message passing, often yield unphysical predictions in these contexts.
These challenges are common in atomistic simulations of molecules and materials, particularly in systems like water, which is ubiquitous in nature and essential to biological processes. 
Moreover, long-range effects are crucial in aqueous solutions, organic electrolytes, and other interfacial systems.

We anticipate that the LES method will be widely adopted in atomistic modeling, especially for systems with significant electrostatic and dielectric properties. 
The long-range MLIP incurs only a modest computational cost, roughly double that of the short-range version (see Methods). 
Furthermore, the LES method can be easily integrated into other MLIP frameworks, such as HDNNP~\cite{behler2007generalized}, ACE~\cite{drautz2019atomic}, GAP~\cite{bartok2010gaussian}, MTPs~\cite{shapeev2016moment}, DeePMD~\cite{zhang2022deep}, NequIP~\cite{batzner20223}, MACE~\cite{batatia2022mace}, etc.

\section{Methods}
\subsection{Implementation}
We implemented the LES method as an Ewald module in CACE, which is written using PyTorch. The code is available in \url{https://github.com/BingqingCheng/cace}.
The current implementation of the Ewald summation should in principle follow an overall scaling of $\mathcal{O}(N^{3/2})$~\cite{toukmaji1996ewald}, \rev{with $N$ being the number of atoms.}
The implementation of the code can be further optimized using algorithms with $\mathcal{O}(N \log(N))$ scaling \rev{employing techniques with the
charges interpolated to a 3D grid such as the Particle-Particle Particle-Mesh method}~\cite{toukmaji1996ewald}.
The MD simulations can be performed in the ASE package, but the LAMMPS interface implementation is currently absent.

\subsection{Selection of the smearing factor $\sigma$}
\rev{
For a system of point charges, the Ewald summation expression in Eqn.~\eqref{eq:ewald} implies that the long-range energy corresponds to the electrostatic energy of Gaussian distributed charges with $\sigma$ being the standard deviation,
while the short-range term needs to account for the difference between the Gaussian charges and the point charges.
The value of the smearing factor $\sigma$ needs to be a fraction of the real-space cutoff radius $r_\mathrm{rut}$ in order to effectively eliminate the short-range error~\cite{kolafa1992cutoff}.
Meanwhile, a larger $\sigma$ makes the Ewald series in $\mathbf{k}$ (Eqn.~\eqref{eq:ewald}) more converged and avoids a high reciprocal space cutoff $k_c$.
and its optimal choice has been extensively discussed in the literature~\cite{kolafa1992cutoff}. 
Combining these considerations, $\sigma$ values in the rough range of 0.5~\AA{} and 2~\AA{} are reasonable choices.

We tested the choice of $\sigma$ on the bulk water dataset, with same settings as before and no message passing layer.  
Fig.~\ref{fig:water-sigma} shows that $\sigma=1$~\AA{} in this case leads to the lowest error, although the differences in outcome between various $\sigma$ choices are quite small.

\begin{figure}
  \centering
  \includegraphics[width=0.45\textwidth]{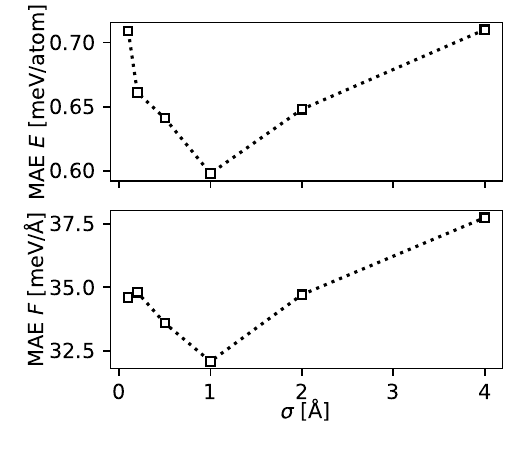}
    \caption{
Energy ($E$) and force ($F$) mean absolute errors (MAEs) on the bulk water dataset~\cite{cheng2019ab} with difference values of the smearing factor $\sigma$,
using short-range models without  massage passing.
    }
    \label{fig:water-sigma}
\end{figure}
}

\subsection{Benchmark of inference speed}
\begin{figure}[h]
  \centering
\includegraphics[width=0.45\textwidth]{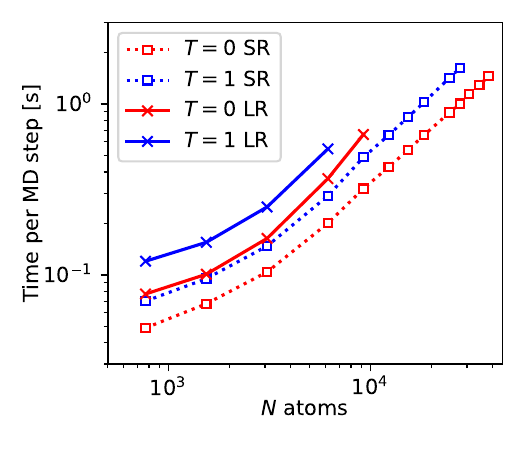}
    \caption{
Timing of molecular dynamics simulations of the bulk liquid water system \rev{with varying number of atoms ($N$) on a L40S GPU,
using short-range (SR) or long-range (LR) CACE models and with or without massage passing layers ($T=0$ or $T=1$). }
    }
    \label{fig:timing}
\end{figure}

We benchmarked short- and long-range CACE water models with and without message passing for MD simulations of liquid water using a single Nvidia L40S GPU with 48 GB of memory. 
Fig.~\ref{fig:timing} illustrates the time required per MD step. All models, with and without LR or message passing, exhibit \rev{favorable} scaling. SR models support simulations with up to approximately 30,000 atoms on a single GPU, while LR models handle around 10,000 atoms due to the higher memory demands of the Ewald routine, which may still be optimized. 
LR models run at roughly half the speed of SR models.
It may be sufficient to use a lower $\mathbf{k}$-point cutoff in the LR model to further increase the speed.

\subsection{MD of Bulk Water}
For each MLIP, we performed NVT simulations of bulk water at 1~g/mL and 300~K to compute the RDF shown in Fig.~\ref{fig:water-gor}. 
The simulation cell contained 512 water molecules, with a time step of 1 femtosecond, using the Nosé-Hoover thermostat. 
The total simulation time was 300~ps. 

To compute the longitudinal component of the dipole density correlation function for each model, we performed an NVT simulation of 2,048 water molecules at 1~g/mL and 300~K in an elongated simulation box with a $z$-dimension of 99.3~\AA{}. 
The simulation time for this system was 200~ps.

To compute the dipole density correlation functions, 
we calculated the dipole moment of each water molecule assuming that hydrogen atoms and oxygen atoms have charges of $+0.4238e$ and $-0.8476e$, respectively. This assumption only affects the absolute amplitude of the correlation function in Fig.~\ref{fig:water-dc}, without altering the relative scale.

\subsection{MD of Water Liquid-Vapor Interfaces}
For the NVT simulations of the thinner slab at 300~K, we used a simulation cell with dimensions $(25.6~\mathrm{\AA}, 25.6~\mathrm{\AA}, 65.0~\mathrm{\AA})$, containing 522 water molecules. 
The thicker slab system had dimensions $(24.8~\mathrm{\AA}, 24.8~\mathrm{\AA}, 120.0~\mathrm{\AA})$, containing 1,024 water molecules. 
In both cases, the width of the water slab was less than half the total width of the cell. 
For the thinner slab, each independent NVT simulation lasted for about 400~ps.
One run using $T=1$ SR model became unstable after about 280~ps
and we discarded the unstable portion.
For the thicker slab, each run lasted for about 600~ps. 
One simulation using the $T=1$ SR model became unstable after about 250~ps, and we only used the stable portion for the analysis.

\textbf{Data availability}

The training scripts, trained CACE potentials, and MD input files are available at \url{https://github.com/BingqingCheng/cace-lr-fit}.

\textbf{Code availability}
The CACE package is publicly available at \url{https://github.com/BingqingCheng/cace}.

\textbf{Acknowledgements}

BC thanks David Limmer for providing the water slab dataset,
and Carolin Faller for the NaCl dataset.

\textbf{Competing Interests}
The author declares no Competing Non-Financial Interests but the following Competing Financial Interests: BC has an equity stake in AIMATX Inc.

\textbf{Author Contributions}
BC designed and performed the study, and wrote the paper.

\end{document}